\pdfoutput=1

\documentclass[11pt]{article}

\usepackage[]{naacl2021}

\usepackage{times}
\usepackage{latexsym}

\usepackage[T1]{fontenc}

\usepackage[utf8]{inputenc}

\usepackage{microtype}
\usepackage{amsmath}
\usepackage{graphicx}
\usepackage{caption}
\usepackage{subcaption}
\usepackage{multirow}
\usepackage{amsfonts}
\usepackage{makecell}
\graphicspath{ {./figures/} }
\title{Cross-Lingual BERT Contextual Embedding Space Mapping with Isotropic and Isometric Conditions}

\author{Haoran Xu \and Philipp Koehn\\
Johns Hopkins University\\
\texttt{hxu64@jhu.edu, phi@jhu.edu}\\
}
\date{}

\begin{document}
\maketitle
\begin{abstract}
Typically, a linearly orthogonal transformation mapping is learned by aligning static type-level embeddings to build a shared semantic space. In view of the analysis that contextual embeddings contain richer semantic features, we investigate a context-aware and dictionary-free mapping approach by leveraging parallel corpora. We illustrate that our contextual embedding space mapping significantly outperforms previous multilingual word embedding methods on the bilingual dictionary induction (BDI) task by providing a higher degree of isomorphism. To improve the quality of mapping, we also explore sense-level embeddings that are split from type-level representations, which can align spaces in a finer resolution and yield more precise mapping. Moreover, we reveal that contextual embedding spaces suffer from their natural properties --- anisotropy and anisometry. To mitigate these two problems, we introduce the iterative normalization algorithm as an imperative preprocessing step. Our findings unfold the tight relationship between isotropy, isometry, and isomorphism in normalized contextual embedding spaces \footnote{Code is available at: \url{https://github.com/fe1ixxu/Contextual_Mapping}.}.
\end{abstract}
\section{Introduction}
\citet{mikolov2013exploiting} first notice that the word vectors pretrained by monolingual data have similar topology structure in different languages, which allows word embedding spaces to be aligned by a simple linear mapping. Orthogonal mapping  \citep{xing-EtAl:2015:NAACL-HLT} is subsequently proved to be an effective improvement for space alignment. With the development of multilingual tasks, \textbf{c}ross-\textbf{l}ingual \textbf{w}ord \textbf{e}mbeddings (CLWE) have attracted a lot of attention in recent times. CLWE facilitates model transfer between languages by providing a shared embedding space, where vector representations of words which have similar meanings from various languages are spatially close. Previous methods can be basically classified into two categories: optimizing a linear transformation to map pretrained word embedding vectors \citep{mikolov2013exploiting,xing-EtAl:2015:NAACL-HLT,artetxe-etal-2016-learning,zhang-etal-2019-girls}, and jointly learning word representation for multiple languages \citep{luong-etal-2015-bilingual,gouws2015bilbowa}. Some recent studies even alternatively work out transformation by unsupervised learning \citep{miceli-barone-2016-towards,zhang-etal-2017-adversarial,lample2018word}.

In this paper, we focus on supervised linear mapping methods. Linear mappings with orthogonal constraints are in the light of the assumption that monolingual word embedding graphs are ( approximately) isomorphic across different languages \citep{sogaard-etal-2018-limitations}. However, this assumption is also a significant limitation because the different structural properties (e.g., morphology, syntax) across languages make it difficult for static word embeddings to meet the hypothesis. 

Instead of utilizing static word embeddings  \citep{mikolov2013efficient,pennington2014glove,bojanowski-etal-2017-enriching},
\citet{schuster-etal-2019-cross} try leveraging word embeddings extracted from ELMo  \citep{peters-etal-2018-deep} to align the embedding spaces by using gold dictionaries, and show the advantages in existing transfer tasks. 

\begin{figure}
     \centering
     \begin{subfigure}[b]{0.46\textwidth}
         \centering
         \includegraphics[width=\textwidth]{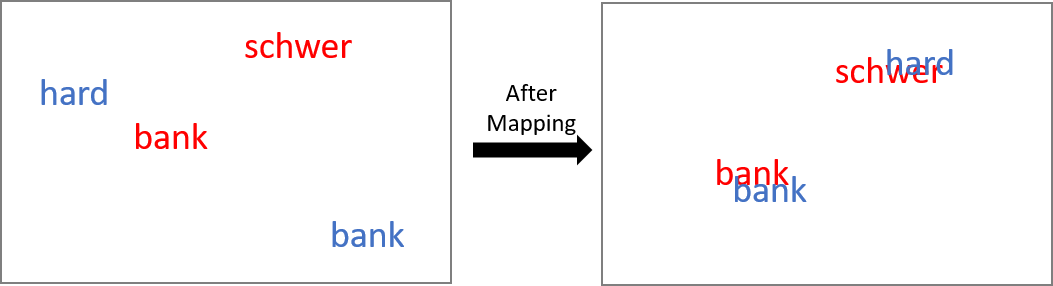}
         \caption{Type-level mapping: English words are accurately mapped to their German translation words. }
         \label{fig:wordmap}
     \end{subfigure}
     \hfill
     \begin{subfigure}[b]{0.46\textwidth}
         \centering
         \includegraphics[width=\textwidth]{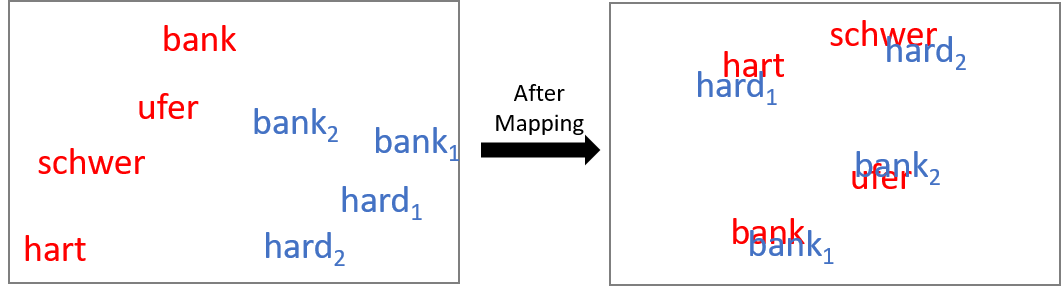}
         \caption{Sense-level mapping: `bank' is split into two sense embeddings, i.e., $\text{bank}_1$ for the meaning of financial establishment and $\text{bank}_2$ for the meaning of shore, which are respectively mapped to German `bank' and `ufer'. A similar discussion also holds for the word `hard', where its two sense vectors are mapped to German `schwer' (difficult) and `hart' (solid). }
         \label{fig:sensemap}
     \end{subfigure}
     \caption{Illustration of contextual cross-lingual mapping of English and German, where word and sense vectors are visualized by t-SNE \citep{maaten2008visualizing}.}
     \label{fig:map}
\end{figure}

Since high-quality, freely available, wide-coverage manually dictionaries are still rare \citep{ruder2019survey}, we investigate an alignment approach that builds silver token translation pairs from parallel corpus rather than leverages gold dictionaries. We first obtain aligned contextual type-level embeddings in both source and target sides simultaneously by averaging vectors of all occurrences of the silver aligned word pairs. Furthermore, we adaptively split a type-level representation into several sense-level representations, where each sense vector in the semantic space represents one of the meanings of the word. A visualization of how these fine-grained sense-level vectors are assigned as anchor vectors to assist in aligning embedding spaces is illustrated in Figure \ref{fig:map}. 

We also explore the properties of contextual embeddings. Compared with static type-level embeddings, experimental results show that better isomorphism of contextual embeddings is the main reason to explain the superior performance of our contextual-aware mapping on the BDI task. Moreover, sense-level embeddings are demonstrated to have a closer isomorphic structure than type-level embeddings. Interestingly, we also discover that contextual embeddings suffer from the problems of anisotropy \citep{ethayarajh2019contextual} and anisometry. Anisotropy is an inherent problem of embedding vectors, where the direction of vectors in the semantic space are not uniformly distributed. Vectors from different languages possess various degrees of anisotropy, which deteriorates the performance of the mapping. Anisometry is also a factor of misalignment because orthogonal mapping is a distance-preserving projection, which implicitly adds an additional restriction in the aforementioned assumption of isomorphism, that is \textbf{isometrical} isomorphism. However, relative distances across languages are usually different (anisometric). To tackle these problems, we introduce the iterative normalization method \citep{zhang-etal-2019-girls}, and show the importance of isotropy and isometry, which effectively improve the quality of mapping.

\section{Background}

\subsection{Supervised Bilingual Space Alignment}
\citet{mikolov2013exploiting} take the lead in exploiting the topological similarities of monolingual embedding spaces to successfully learn a linear mapping. Although more complicated models like multi-layer neural networks are tried to implement on space alignment, they don't observe any improvements. To bridge a link between source and target language spaces, a gold dictionary has to be used. Let denote $\{x_i,y_i\}_{i\in\{1,\cdots,n\}}$ as word embedding vectors corresponding to the $n$ translation pairs in the dictionary. A linear matrix $\hat{W}$ is learned by minimizing the Frobenius norm:

\begin{equation}
    \label{min}
    \hat{W} = \mathop{\arg\min}_{W\in \mathbb{R}^{d\times d}}\|WX-Y\|_F
\end{equation}
where $X$ and $Y$ are embedding matrix composed of embedding vectors of word pairs in the dictionary, and $d$ is the dimension of embedding vectors. Since the objective function is convex, $\hat{W}$ can be solved by gradient-based methods. Alternatively, \citet{xing-EtAl:2015:NAACL-HLT} shows better results of bilingual space alignment by enforcing orthogonal limitation on linear mapping, so the optimization function boils down to the Procrustes problem \citep{schonemann1966generalized}, which offers a close form solution:
\begin{equation}
    \label{procruste}
    \hat{W} = \mathop{\arg\min}_{W\in \mathbb{O}^{d\times d}}\|WX-Y\|_F = UV^T
\end{equation}
where $U\Sigma V^T = \text{svd}(YX^T)$, and $\mathbb{O}^{d\times d}$ is a set of orthogonal matrices with size $d\times d$.

\subsection{Hubness Problem}
Word retrieval task is a metric to evaluate the quality of embedding space mapping. Still, it is known to suffer from the hubness problem \citep{radovanovic2010hubs}, where a few vectors (hubs) in high dimension spaces are the nearest neighbors of many other vectors. This phenomenon undermines the performance of retrieval methods which are established on the nearest neighbor rules. In lieu of matching pairs by finding the nearest neighbour vector, \citet{lample2018word} propose \textbf{c}ross-domain \textbf{s}imilarity \textbf{l}ocal \textbf{s}caling (CSLS) criterion to alleviate the hubness problem by penalizing the similarity score of hubs.

\subsection{Quantifying Isomorphism}
\label{sec:rs}
The degree of isomorphism reflects how topologically similar the structures of the two vector spaces are. A higher degree of isomorphism across two vector spaces always means that they are easier mapped by an orthogonal matrix. Based on the analysis that if two semantic language spaces possess a high degree of isomorphism, the similarity distribution of words in the same meaning within each language should be similar, \citet{vulic2020all} propose \textbf{r}elational \textbf{s}imilarity (RS) metric. Formally speaking, $M$ translation pairs are extracted from bilingual embeddings first. Then, $\forall i,j\in\{1,\cdots,M\}, i\neq j$, we list the similarity of all word pairs $(w_i^s,w_j^s)$, where $w_{i}^s, w_j^s$ are $i_{th}$ and $j_{th}$ word in the source side, and analogously, obtain a similar list in the target side. Finally, Pearson correlation coefﬁcient $\rho$ of the two lists is calculated to evaluate the degree of isomorphism. The correlation coefficient increases with a higher degree of isomorphism. It is worth mentioning that $\rho = 1$ if two embedding spaces are isomorphic.

\subsection{Contextual Word Embeddings}
The remarkable progress of NLP tasks utilizing pretrained monolingual models \citep{peters-etal-2018-deep,devlin-etal-2019-bert,yang2019xlnet} illustrates the crucialness of contextual representation. One straightforward way to obtain contextual type-level embeddings is averaging the vectors of words from the monolingual corpus fed into the pretrained language model.  An offline transformation matrix can be learned by these contextual vectors and has been successfully applied into a zero-shot cross-lingual dependency parsing task \citep{schuster-etal-2019-cross}. However, gold dictionaries are still necessary prerequisites to align the spaces, and words in a different context cannot always be accurately translated by static dictionaries. Importantly, type-level representations for multi-sense words are biased because they tend to express their majority meaning, potentially harming the space alignment (more details in Section \ref{sec:representation_bias}). 

The predominant performance of pretrained models also attracts attention to the hierarchy of linguistic information, where an in-depth study of BERT \citep{devlin-etal-2019-bert} is conducted by \citet{jawahar-etal-2019-bert}. They reveal that surface features, syntactic features, and semantic features of words respectively lie on the bottom layers, middle layers, and top layers of BERT. In this paper, the representations of words used for aligning the bilingual spaces are the normalized mean vectors of the topmost four layers of pretrained BERTs.

\section{Approach}
We next describe the context-aware embedding mapping approaches and properties of embeddings.
\subsection{Preprocessing}
\textit{Fast Align} \citep{dyer-etal-2013-simple} is a log-linear reparameterization of IBM Model 2 \citep{Brown:1993}, which is an effective unsupervised bidirectional token alignment algorithm. Instead of using a dictionary to derive translation pairs, we apply \textit{Fast Align} to parallel corpora to obtain the silver aligned token pairs. This offers us three advantages for mapping over using a dictionary: 1) parallel corpora provide a more comprehensive range of scope for translation pairs than a dictionary; 2) embeddings of translation token pairs contain the same contextual information; 3) tokens are aligned in each parallel sentence, and embeddings are already aligned as well, so mappings can be created by aligning embeddings and skip the step of word alignment from a dictionary.

\subsection{Contextual Type-Level Embeddings Alignment }\label{cwea}
A tokenized parallel corpus is fed into pretrained BERTs \citep{Wolf2019HuggingFacesTS,safaya2020kuisail} of the source and target languages. Since every occurrence of a token possesses a contextual word embedding and a type commonly appears multiple times in the corpus, we always receive a collection of contextual vectors for a type. On the source side, type-level representation $x_i$ of a type $i$ is defined as the mean vector of all vectors in its collection. Because \textit{Fast Align} bridges a link between token pairs in every parallel sentence and embeddings of the token pairs, the mean vector $y_i$ of linked target vectors can be simultaneously derived. Then, we build two column-wise aligned embedding matrices $X$, $Y$, where $X = [x_1, \cdots, x_N]$, $Y = [y_1, \cdots, y_N]$ and $N$ is the vocabulary size of the source language. We derive the optimal orthogonal transformation $\hat{W}$ by leveraging the close form solution in Equation \ref{procruste}. 


\subsection{Contextual Sense-Level Embeddings Alignment}
\begin{figure}[t]
    \centering
    \includegraphics[width=7.5cm]{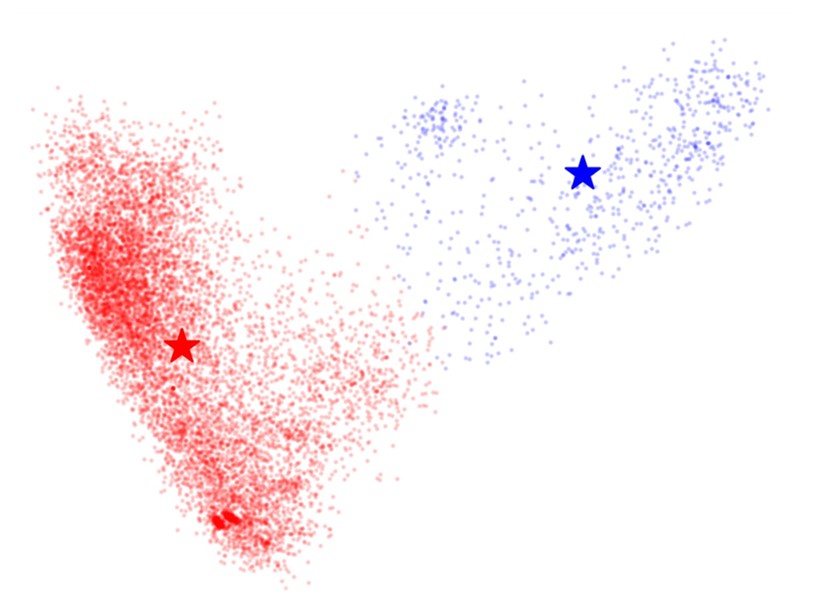}
    \caption{The distribution of contextual embedding vectors of word `bank' in 2 dimension PCA, where the red points in the left side represent the financial meaning in their contexts, while blue points in the right side are the meanings of `shore'. The center of the clusters are illustrated as stars.}
    \label{fig:bank}
\end{figure}

Unlike type-level alignment, we align the space in a finer resolution by leveraging sense vectors, where each meaning of a word has its representation.
\paragraph{Well Separated Contextual Vectors:}
Intuitively, contextual vectors of a word with different meanings are expected to be distributed in distinct locations of semantic space, and vectors that represent the same meanings should be tightly adjoined together. A visualized example of contextual vectors of a word `bank' is given in Figure \ref{fig:bank}. In this case, two clusters corresponding to different meanings of `bank' are spatially opposite to each other. This observation supports the anchor-driven mapping, which is realized by aligning the centers of clusters --- sense-level representation.

\paragraph{Sense-Level Representation:}
We cluster the embedding vectors in the source side by $k$-means algorithm and define the mean vector of a cluster as a sense-level embedding vector. The algorithm of finding out the optimal number $k$ of clusters follows an elbow-based approach \citep{satopaa2011finding} \footnote{We find that the performance of adaptively detecting $k$ is better than setting $k$ as a constant number.}, where the `knee' point \footnote{We also have tried Gap statistic algorithm \citep{tibshirani2001estimating}. However, it is more time-expansive, and our preliminary experiments show that its performance is weaker than the knee detection approach.} at which the optimal $k$ locates can be adaptively detected. Importantly, this approach refuses to cluster a vector collection that is adjudicated as not well separate, which promises that mono-sense words are not over-clustered.

\paragraph{Space Alignment:}
For each type $i$, a list of sense-level representations $x_{i,1},\cdots, x_{i,|s_i|}$ is derived after clustering contextual vectors, where $|s_i|$ is the number of clusters. Similar to the alignment in Section \ref{cwea}, each source embedding is linked to a target embedding, so a sense-level embedding $x_{i,j}$ and the corresponding target embedding $y_{i,j}$ can be obtained in the meantime, where integer $j\in \{1,\cdots,|s_i|\}$. Two sense-level aligned matrices $X^s$, $Y^s$ are generated by concatenating all their sense-level vectors. Namely, $X^s=[\cdots, x_{\cdot,1},\cdots, x_{\cdot,|s_\cdot|}, \cdots]$, and $Y^s=[\cdots, y_{\cdot,1},\cdots, y_{\cdot,|s_\cdot|}, \cdots]$, where $\cdot$ denotes all possible types in the source vocabulary. Finally, we derive an optimal orthogonal transformation by Equation \ref{procruste}.

\paragraph{Intuition behind Sense-Level Embeddings --- Solving Representation Bias Problem:}
\label{sec:representation_bias}
Contextual type-level embeddings have one apparent phenomenon that they are inclined to embody their majority meanings, i.e., the most frequent meanings that appear in the corpus. In other words, type-level embedding vectors of multi-sense words tend to be closer to the vectors representing their primary meaning in the semantic space. This phenomenon brings up a drawback that multi-sense word vectors are difficult to accurately represent any of their senses. The representation bias for multi-sense words potentially degenerate the quality of embedding vectors and deteriorate the accuracy of cross-lingual mapping. To mitigate the representation bias problem, we investigate the sense-level representation. 

\begin{figure}
     \centering
     \begin{subfigure}[b]{0.23\textwidth}
         \centering
         \includegraphics[width=\textwidth]{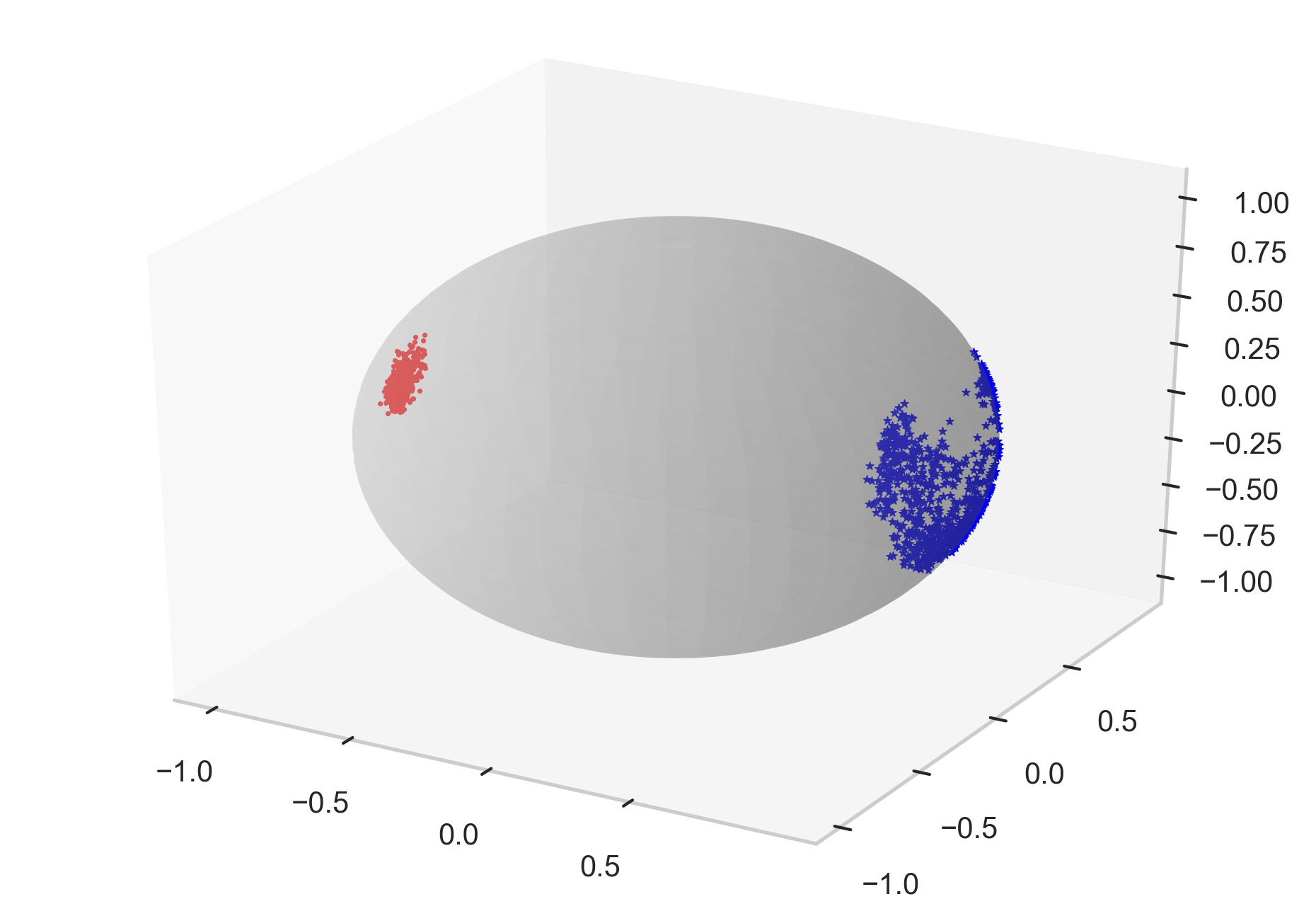}
         \caption{Before iterative normalization: monolingual vectors only gather in a small area. }
         \label{fig:anisotropy}
     \end{subfigure}
     \hfill
     \begin{subfigure}[b]{0.23\textwidth}
         \centering
         \includegraphics[width=\textwidth]{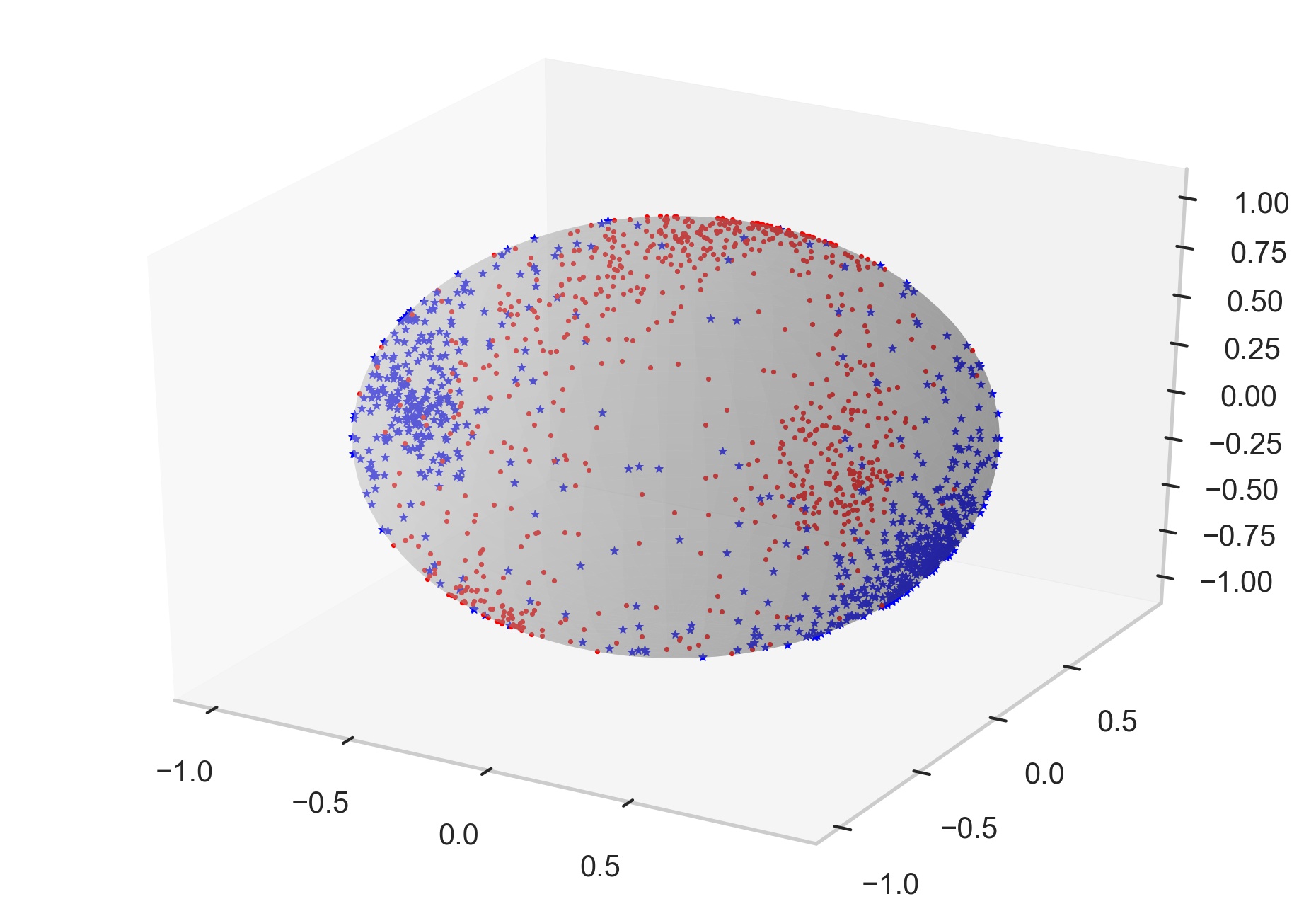}
         \caption{After iterative normalization: monolingual vectors are uniformly distributed.}
         \label{fig:isotropy}
     \end{subfigure}
     \caption{The distribution of normalized contextual word embeddings on the surface of a sphere, where blue stars are English word vectors and red points are German word vectors. 3-d vectors are derived by PCA dimension reduction.}
\end{figure}

\subsection{Properties of Embedding Spaces}
Here we introduce two important concepts: \textbf{isotropy} and \textbf{isometry}, and reveal their influence on improving the degree of isomorphism for contextual spaces, which offers higher-quality mapping. Our findings show that \textbf{isomorphism is positively correlated with isometry, and isometric spaces can be built by enforcing spaces to be isotropic.}

\paragraph{Isotropy:} An embedding space is isotropic if the directions of embedding vectors are uniformly distributed. Unfortunately, contextual word representation is usually anisotropic. Geometric speaking, normalized word embedding vectors are more likely to gather on a narrow conical surface of a hypersphere rather than uniformly distributed in all directions. Commonly, various language vector spaces have various degrees of anisotropy. Figure \ref{fig:anisotropy} illustrates the different size of areas that contextual English and German vectors occupy, where vectors are obtained from a parallel corpus. We give a simple metric to evaluate the degree of isotropy:

\begin{equation}
    \label{isotropy}
    D^t = \frac{2}{r(r-1)}\sum_{i,j, i\neq j} v_i^Tv_j
\end{equation}
where $v_i,v_j$ are normalized vectors in the space, $r$ is the number of randomly selected vectors, and most importantly, $D^t$ represents the average cosine similarity between all selected normalized vectors. The closer the average similarity is to 0, the more isotropic the embedding space is. An isotropic space should hold the property that $D^t$ equals 0.

\paragraph{Isometry:}
\label{sec:isometry}
We define two spaces are isometric if relative Euclidean distances among vectors are identical between spaces. Orthogonal mapping is a distance-persevering and isometrically isomorphic transformation so that Euclidean distance between two vectors does not change after mapping. Thus, two semantic language spaces are more comfortable to be aligned if their relative distances of vectors are similar. We measure the degree of isometry between two embedding spaces by calculating the average of absolute difference of relative distances:
\begin{equation}
    \label{isometry}
    D^m=\frac{2}{r(r-1)}\sum_{i,j, i\neq j}\Big |\|x_i-x_j\|_2^2
-\|y_i-y_j\|_2^2 \Big |
\end{equation}
where $x_i,x_j$ are vectors in the source space and $y_i, y_j$ are the translation vectors in the target space. The lower $D^m$ is, the more isometric the bilingual spaces are. Two spaces are isometric when $D^m$ is $0$. Since vectors in embedding spaces are normalized, Equation \ref{isometry} boils down to:
\begin{equation}
    \label{isometry-boil-down}    
    D^m=\frac{4}{r(r-1)}\sum_{i,j, i\neq j}\Big |x_i^Tx_j - y_i^Ty_j \Big |
\end{equation}

\paragraph{Iterative normalization:}
\label{sec:IN}
Anisometry undermines the quality of mapping due to the inconsistent relative distances of embeddings in the source and target embedding spaces. Therefore, we look for a way to reduce (increase) the degree of anisometry (isometry) to mitigate its negative influence. Relying on Equation \ref{isotropy} and \ref{isometry-boil-down}, two semantic spaces are near-isometric when they have a similar degree of isotropy. However, it is unrealistic to control spaces in the same attribute (an)isotropic degree, but alternatively, we introduce a preprocessing method, namely \textbf{i}terative \textbf{n}ormalization (IN). We apply it to transform anisotropic contextual embedding spaces to be (approximately) isotropic ($D^t\approx0$) by forcibly distributing vectors uniformly on the surface of the unit hypersphere (Figure \ref{fig:isotropy}). This highly improves the degree of isometry, i.e., relative embedding distances across language spaces are more similar. The preprocessing method iteratively enforces vectors to be zero-mean and be normalized. We let $x_i^0$ denote the initial embedding for word (sense) $i$. For every word (sense), embedding vectors are firstly enforced to be normalized to a unit length in the $k^{th}$ iteration:
\begin{equation}
    z_i^k = \frac{x_i^{k-1}}{\|x_i^{k-1}\|}
\end{equation}
and make it be zero-mean:
\begin{equation}
    x_i^{k} = z_i^k - \frac{1}{N}\sum_{i=1}^Nz_i^k
\end{equation}
where $N$ is the number of words (senses). We repeat the two steps above until convergence.  

\section{Experiment}
We evaluate the cross-lingual contextual mapping between English and three other languages --- German (de), Arabic (ar), and Dutch (nl), where German and Dutch are closely related to English, and Arabic is distinct to English. We set English as the source language and other languages as targets.

\subsection{Settings}
\paragraph{Dataset:}
Parallel corpora of European languages are downloaded from \emph{ParaCrawl v6.0} \footnote{\url{www.paracrawl.eu}}, while the English-Arabic parallel corpus is extracted from \emph{United Nations Parallel Corpus} \footnote{\url{https://conferences.unite.un.org/uncorpus/}}. We select 500K parallel sentences for each language pairs and truncate sentences whose length is longer than 150.

\paragraph{Token Alignment:}
Accurate alignment is not always offered by \textit{Fast Align}, so we do not take one-to-many, many-to-one, and many-to-many alignments into consideration. We do not use subword embeddings to approximately represent an \textbf{O}ut-\textbf{o}f-\textbf{v}ocabulary (OOV) token, as aligned embeddings of non-OOV tokens are easier to have more similar relative positions across embedding spaces.  

\paragraph{Contextual Embeddings Extraction:}
For types in the vocabulary that appear at least five times in the given parallel corpus, we store at most 10K contextual vectors. When we extract sense-level embeddings, we only cluster types whose occurrence in the corpus is over 100. Sense-level embedding vectors for \emph{stopwords} \footnote{We use the lists of stopwords from \url{https://github.com/Alir3z4/stop-words}} and low-frequent types (occurrence under 100) are just obtained by averaging all their vectors --- they boil down to type-level representations. Note that contextual embeddings are normalized to a unit length. BERT models of all languages are base-size (12 layers with 768 dimensions). 

\subsection{Bilingual Dictionary Induction}
\begin{table*}[h]
\begin{center}

\begin{small}
  \begin{tabular}{cc|cccccccccccc}
 \hline
     && \multicolumn{2}{c}{\bf de} && \multicolumn{2}{c}{\bf ar} && \multicolumn{2}{c}{\bf nl} \\
    \textit{Before iterative normalization} && P@1 & P@5 && P@1 & P@5 && P@1 & P@5 \\
    \hline
    FastText - NN && 72.80 & 87.22 &&  52.13 &  76.94 && 73.23 & 90.21  \\
    Ours, type-level - NN  && \bf 81.60 & 95.30 && \bf 71.93 & 95.74 && 86.36 & 97.17 \\
    Ours, sense-level - NN && \bf 81.60 & \bf 96.42 &&  71.43 & \bf 96.00 && \bf 87.00 & \bf 97.30  \\
    \hline
    FastText - CSLS && 77.71 & 88.96 &&  64.91 &  81.95 && 79.02 & 91.63  \\
    Ours, type-level - CSLS  && 85.28 & 96.73 && 79.70 & 95.99 && 89.32 & \bf 98.33 \\
    Ours, sense-level - CSLS && \bf 85.58  & \bf 97.65 && \bf 80.70 & \bf 96.24 && \bf 89.70 & 97.81  \\
   \hline
    \hline
    \textit{After iterative normalization} && \multicolumn{2}{c}{} && \multicolumn{2}{c}{} && \multicolumn{2}{c}{} \\
    \hline
    FastText - NN && 73.01 &  87.93 && 53.88 &  77.70 && 75.16 & 90.86  \\
    Ours, type-level - NN  && 81.60 & 95.81 && 72.43 & \bf 96.49 && \bf 86.10* & 97.17 \\
    Ours, sense-level - NN && \bf 82.41 & \bf 96.83 && \bf 74.44 &  96.00 && \bf 86.10* & \bf 98.01  \\
    \hline
    FastText - CSLS && 77.61* & 89.37 &&  64.16* &  82.96 && 79.79 & 92.54  \\
    Ours, type-level - CSLS  && 86.30 & 97.03 && 80.70 & \bf 96.24 && 91.12 & 98.46 \\
    Ours, sense-level - CSLS && \bf 87.12 & \bf 98.16 && \bf 81.20 & \bf 96.24 && \bf 91.51 & \bf 98.58  \\
   \hline
  \end{tabular}
\end{small}
\end{center}
\caption{Evaluation measures for the three cross-lingual embedding mapping approaches before and after iterative normalization. (*) means that the score does not increase after iteartive normalization. `FastText' in the table refers to the supervised mapping from \citet{lample2018word}.}
\label{tab:results:precision}
\end{table*}
We evaluate our mappings on the BDI task, which considers a problem of a target language word retrieval for each query source word relying on the representation in the space after a shared cross-lingual embedding space is built. 

\paragraph{Baselines:}
Our main baseline is supervised mapping \citep{lample2018word} that uses Procrustes solution for fastText embeddings \citep{bojanowski-etal-2017-enriching}. The second baseline is supervised mapping for ELMo embeddings from \citet{schuster-etal-2019-cross}. Following their experimental settings, the outputs of the first LSTM layer are used for representing tokens. For a fair comparison, the anchor vectors are derived by utilizing the same corpora as what we use to generate our contextual embeddings. We only compare our method with the ELMo embedding mapping in English-German language pair for which they support the off-the-shelf monolingual pretrained ELMo models \footnote{They do not support Dutch and Arabic ELMo models.}.

\paragraph{Training and Evaluation:}
Although our approach is dictionary-free, mapping methods from \citet{lample2018word} and \citet{schuster-etal-2019-cross} need a seed dictionary. To fairly reveal the comparison of our mappings and the baselines, we use identical training pairs in the same seed dictionary and evaluate mappings on the identical translation pairs in the test dataset. Mappings are obtained by leveraging the 5k most frequent words in the source language and their translations in the dictionaries. Dictionaries take into account the polysemy of words and are publicly available in the MUSE library \footnote{\url{https://github.com/facebookresearch/MUSE}}. We evaluate mappings based on 1.5k word pairs whose source words rank in frequency from 5K to 6.5K, where source word queries use 200K target words. Note that we do not consider the OOV tokens, so the training and test pairs are the overlapping tokens in the dictionary and non-subwords in the BERT vocabulary. Also, queries use the overlapping tokens of 200K target words and non-subwords in the BERT vocabulary. Thus, for en-de, en-nl, and en-ar, we use 4191, 3375, 2605 training pairs, 978, 777, 399 source test queries, and 14140, 12352, 7918 target words, respectively. We retrieve target words by finding the nearest neighbor (largest cosine similarity) to the source words and then repeat the retrieval task again by using the CSLS metric with the setting number of neighbors as 10.
 

\subsection{Degree of Isomorphism}
To illustrate the close relationship between isotropy, isometry, and isomorphism, we also compute their scores. For each language pairs, we leverage the aforementioned relational similarity (RS) in Section \ref{sec:rs} to measure the degree of isomorphism. We use $M=1500$ translation pairs whose source words are most frequent to calculate RS. The number of randomly selected vectors is 1K for the calculation of isotropic degree $D^t$ and isometric degree $D^m$.

\subsection{Iterative Normalization}
For all experiments above, we conduct experiments both with and without iterative normalization. Experiments run for 5 iterations, which is sufficient to converge.

\section{Discussion}
\subsection{Effect of Our Context-Aware Embeddings}
The main results are shown in Table \ref{tab:results:precision}. It indicates that the performance of our contextual embeddings are significantly superior than of static fastText embeddings in the BDI task, where our mappings outperform fastText embedding mappings by approximate 10\% accuracy. Especially in the evaluation of distinct language pairs, English and Arabic, it shows us the most impressive improvement, which boosts the accuracy of P@1 and P@5 almost 20\% higher. Even though the target words for queries are a subset of 200K words, we are surprised that contextual embedding mapping outperforms fastText embedding mapping by a large margin in the same settings of training and test word pairs. The success of contextual mapping is favored with the similar relative positions of our aligned contextual embeddings because they share the same contextual information from the parallel corpus, and embeddings of subwords do not approximately represent them, which builds a high degree of isomorphism for the cross-lingual spaces. Recall that a higher RS score implies a higher degree of isomorphism. As illustrated by the RS column in Table \ref{tab:results:rs}, contextual embeddings are able to construct spaces that possess substantially higher RS scores in comparison to fastText.

In Table \ref{tab:results:ELMo}, our contextual mapping also significantly outperforms the ELMo embedding mapping. Note that we still use the same training and test sets as the fastText alignment experiment.

\subsection{Effect of Sense-Level Embeddings}
Contextual sense-level embeddings are derived by splitting multi-sense word embeddings to mitigate the problem of representation bias (Section \ref{sec:representation_bias}) and construct a higher degree of isomorphism spaces across various language pairs.

\begin{table*}[ht]
\begin{center}

\begin{scriptsize}
  \begin{tabular}{c|cccc|cccc|cccc}
 
    & \multicolumn{4}{c}{\bf de} & \multicolumn{4}{c}{\bf ar} & \multicolumn{4}{c}{\bf nl} \\
    \emph{Before IN} & $D_s^t$ & $D_t^t$ & $D^m$ & RS &  $D_s^t$ & $D_t^t$ & $D^m$ & RS & $D_s^t$ & $D_t^t$ & $D^m$ & RS \\
    
    \hline
    fastText & 0.1650 & 0.1681 & 0.1085 & 0.5452 & 0.1706 & 0.1611 & 0.1362 & 0.3751 & 0.1701 & 0.1996 & 0.1161 & 0.5958 \\
    type-level & 0.4496 & 0.7087 & 0.5184 & 0.6998 & 0.4161 & 0.5001 & 0.1888 & 0.6660 & 0.4784 & 0.7545 & 0.5524 & 0.6804 \\
    sense-level & 0.4187 & 0.6793 & 0.5213 & \bf 0.7334 & 0.4000 & 0.4747 & 0.1796 & \bf 0.6912 & 0.4587 &  0.7326 & 0.5480 & \bf 0.7382 \\

    \hline
    \emph{After IN} & & & & & & & & & & & &\\
    \hline
    fastText  & 0.0035 & 0.0022 & 0.1279 & 0.5245 & 0.0040 & 0.0024 & 0.1459 & 0.3994 & 0.0080 & 0.0052 & 0.1193 & 0.5747\\
    type-level & 0.0089 & 0.0062 & 0.1062 & 0.7586 & 0.0115 & 0.0101 & 0.1091 & 0.7970 & 0.0109 & 0.0068 & 0.1035 & 0.7829 \\
    sense-level & 0.0064 & 0.0043 & 0.1082 & \bf 0.7620 & 0.0112 & 0.0108 & 0.1125 & \bf 0.8019 & 0.0085 & 0.0054 & 0.1044 & \bf0.7832 \\
    \hline
  \end{tabular}
\end{scriptsize}
\end{center}
\caption{Scores of isotropy, isometry and relational simiarlity across embedding spaces before and after \textbf{i}teartive \textbf{n}ormlization (IN) between English and three target languages, where $D_s^t$ and $D_t^t$ are the scores of isotropy in the source and target space respectively. Bold numbers are the highest score of RS among three mapping methods.}
\label{tab:results:rs}
\end{table*}

\begin{table}[h]
\begin{center}

\begin{scriptsize}
  \begin{tabular}{c|ccccc}
 \hline
    & P@1 & $D_s^t$ & $D_t^t$ & $D_m$ & RS \\
    \hline
    ELMo (before \textit{IN}) & 55.73 & 0.2160 & 0.3200 & 0.2425 & 0.4984 \\
    ELMo (after \textit{IN}) & 56.75 & 0.0062 & 0.0053 & 0.1787 & 0.5547 \\
    \makecell[c]{Ours, sense-level \\ (after \textit{IN})}& \bf 87.12 & 0.0064 & 0.0043 &  0.1082 & \bf 0.7620\\
   \hline
  \end{tabular}
\end{scriptsize}
\end{center}
\caption{The comparison of our best mapping with ELMo embedding mapping for English and German. Note that the P@1 scores are derived by CSLS metric.}
\label{tab:results:ELMo}
\end{table}

\paragraph{Mitigation of Representation Bias:}
We select two English words, `bank' and `hard', to expose how sense-level embeddings allay the representation bias problem between English and German. The English word `bank' usually has one financial meaning and meaning of land alongside a river. After we obtain a transformation matrix for English and German language pairs, we map English embedding vectors into German space and calculate the cosine similarity between the English word `bank' with its German translation `bank' (financial meaning) and `ufer' (shore meaning). Table \ref{tab:results:cs} shows that the sense-level embedding that represents the financial meaning is closer to the vector of the German word `bank' compared with type-level embeddings, and another sense-level vector is also correctly mapped to the neighbor of `ufer'. Note that `bank' is in the most frequent 5K training pairs, so the mapping used in Table \ref{tab:results:cs} excludes it during training. The same holds for the English word `hard' with its translation of German words `hart' and `schwer'. 

\paragraph{Compare with Type-Level Embeddings:}
In the comparison of scores on the BDI task illustrated by Table \ref{tab:results:precision}, sense-level embeddings outperform the type-level one by around 1\% accuracy. We attribute the better performance of sense-level embeddings to the slightly higher degree of isomorphism, which has been shown by the RS scores in Table \ref{tab:results:rs}. It reveals that the better isomorphic structure benefits from mitigating the representation bias.

\begin{table}
\begin{center}

\begin{small}
  \begin{tabular}{|c|ccc|}
    \hline
     & en word & de word & cosine simiarity \\
     
    \multirow{2}{*}{type-level}
     & bank & bank & 0.6868 \\
     & bank & ufer & 0.1461 \\  
     \hline
     
    \multirow{2}{*}{sense-level}
     & bank$_1$ & bank & \bf 0.6891   \\
     & bank$_2$ & ufer & \bf 0.6481    \\  
    \hline
    \hline
    
    \multirow{2}{*}{type-level}
     & hard & schwer & 0.6775 \\
     & hard & hart & 0.2551 \\
     \hline
     
    \multirow{2}{*}{sense-level}
     & hard$_1$ & schwer & \bf 0.7286   \\
     & hard$_2$ & hart & \bf 0.6755    \\  
    \hline

  \end{tabular}
\end{small}
\end{center}
\caption{Comparison of cosine similarity between example English words (`bank', `hard') and their translation German words after mapping in the type-level and the sense-level. In the case of sense-level method, bank$_1$ means `financial establishment', bank$_2$ means `shore', hard$_1$ means `difficult', and hard$_2$ means `solid'.  }
\label{tab:results:cs}
\end{table}

\subsection{Effect of Isotropy and Isometry}
As indicated in the Table \ref{tab:results:rs}, the original (before iterative normalization) degree of isotropy of contextual embeddings across English ($D^t_s$ column) and other three target languages ($D^t_t$ column) are very different. The isotropic score of English space is around 0.45, while the score of Arabic space is 0.5, and scores of German and Dutch spaces even exceed 0.7, which implies that the target language embeddings gather in narrower `cones' in the semantic space.  

As discussed in Section \ref{sec:isometry}, two semantic spaces have a higher isometric degree when they have similar degrees of (an)isotropy. Results under the row of \textit{After IN} in Table \ref{tab:results:rs} shows the scores of isotropy are all dropped near to 0, and the scores of isometry drop significantly in the meantime for our contextual mapping methods, which indicates the success of manufacturing better isometric condition ($D^m$ column) by leveraging iterative normalization.

RS scores in Table \ref{tab:results:rs} demonstrate that a higher degree of isomorphic space can basically be constructed with a higher isometric degree. In comparison with the P@1 and P@5 results before \textit{IN} in Table \ref{tab:results:precision}, the superior results after \textit{IN}, corresponding to the higher RS scores (after \textit{IN}) in Table \ref{tab:results:rs}, importantly shows the improvement on BDI task benefits from the higher isomorphic and isometric spaces. Note that the above discussion also applies to ELMo contextual embedding alignment (indicated in Table \ref{tab:results:ELMo}).


\section{Conclusion}
In this paper, our contextual embeddings have unfolded their powerful capacity of building high-quality mappings and also illustrated a higher degree of isomorphism across language spaces compared with previous mapping methods. The success of the contextual embeddings provided us a new sight of exploring cross-lingual spaces by extracting parallel information from deep pre-trained language models. Interestingly, contextual sense-level embeddings showed advantages in space mapping by splitting multi-sense word embedding vectors into several sense vectors, which ameliorated the representation bias problem. We have also explored the relationship of isotropy and isometry for cross-lingual embedding spaces and leveraged iterative normalization to keep the consistency of isometry across languages, which improved the current degree of isomorphism again.

Our future work is to apply our contextual embedding mapping method to downstream cross-lingual transfer tasks with a broader range of high-quality aligned embeddings of translation pairs.


\bibliography{anthology,custom}
\bibliographystyle{acl_natbib}




\end{document}